\documentclass[journal]{IEEEtran}
\ifCLASSINFOpdf
\else
\fi
\usepackage[utf8]{inputenc}
\usepackage[T1]{fontenc}
\usepackage{cite}
\usepackage{amsmath,amssymb,amsfonts}
\usepackage{algorithmic}
\usepackage{graphicx}
\usepackage{textcomp}
\usepackage{xcolor}
\usepackage{breqn}
\usepackage{textcase}
\usepackage[export]{adjustbox}
\usepackage{subcaption}

\begin{document}
%
\title{\LARGE \bf Efficient falsification approach for autonomous vehicle validation using a parameter optimisation technique based on Reinforcement Learning}

\author{Dhanoop Karunakaran$^{1}$, Stewart Worrall$^{1}$, Eduardo Nebot$^{1}$
\thanks{$^{1}$D.Karunakaran, S. Worrall,  E. Nebot  are with the Australian Centre for Field Robotics (ACFR) at the University of Sydney (NSW, Australia).
       E-mails: {\tt\small \{d.karunakaran,  s.worrall,  e.nebot\}@acfr.usyd.edu.au}}
}

\maketitle

\begin{abstract}

The widescale deployment of Autonomous Vehicles (AV) appears to be imminent despite many safety challenges that are yet to be resolved. 
It is well-known that there are no universally agreed Verification and Validation (VV) methodologies to guarantee absolute safety, which is crucial for the acceptance of this technology.
The uncertainties in the behaviour of the traffic participants and of the dynamic world cause stochastic reactions in advanced autonomous systems. In addition to that, modern autonomous vehicles will undoubtedly include machine learning and probabilistic techniques. The addition of ML algorithms adds significant complexity to the process for real-world testing when compared to traditional methods.  The most common approach for evaluating system performance is based on large scale real-world data gathering exercises (number of miles travelled), or a test matrix approach. Recently, the research community and industry have started to move towards scenario-based testing as an alternative option that is considered to be a feasible methodology to realise AV use on public roads.  Most research in this area focuses on generating challenging concrete scenarios or test cases to evaluate the system performance by looking at the frequency distribution of extracted parameters as collected from the real-world data. These approaches generally employ Monte-Carlo simulation and importance sampling to generate critical cases.

This paper presents an efficient falsification method to evaluate the System Under Test. The approach is based on a parameter optimisation problem to search for challenging scenarios. The optimisation process aims at finding the challenging case that has maximum return. Our method is based on Neural Architecture Search (NAS) to search for the optimal parameter combinations. NAS applies policy-gradient reinforcement learning algorithm to enable the learning. The riskiness of the scenario is measured by the well established RSS safety metric, euclidean distance, and instance of a collision. We demonstrate that by using the proposed method, we can more efficiently search for challenging scenarios which could cause the system to fail in order to satisfy the safety requirements or evaluation metrics.  
\end{abstract}

\section{Introduction}

A guarantee of safety is essential for the acceptance of autonomous vehicles, and society is unlikely to tolerate fatalities caused by failures of an intelligent system\cite{ITS_2020_workshop}\cite{mobileye_rss}\cite{ITS_2019_workshop}. The validation of Highly Automated Vehicles (HAV) is an essential step before deploying them on public roads. Traditionally, the test matrix approach is considered to be one of the primary validation methods to evaluate the system. It uses test scenarios extracted from a crash database \cite{zhao2016accelerated}. A typical example based on this approach is the Autonomous Emergency Braking (AEB) test protocol developed as part of the Euro New Car Assessment Program (Euro-NCAP)\cite{sarkar2019behavior}. Even though this approach is repeatable and reliable, these scenarios are fixed and predefined\cite{zhao2016accelerated}. As a result, the system is likely to perform well in these tests, but this is not a true test of real-world performance as the real-world is non-deterministic and challenging\cite{zhao2016accelerated}. 

\begin{figure}[t]
\includegraphics[width=0.98\columnwidth]{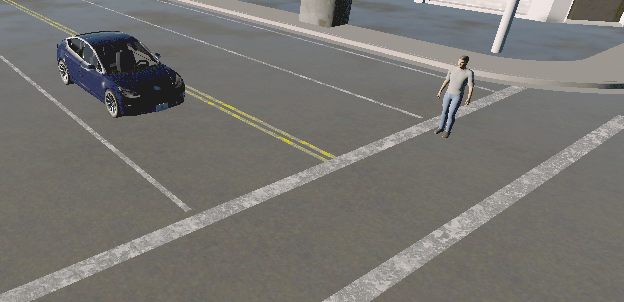}
\caption{\small Carla simulation of the experiment: a pedestrian crossing the street while the vehicle is moving. }
\label{fig:ped_cross_sample}
\end{figure}

One possible method to evaluate the system is to use a Naturalistic Field Operation Test (N-FOT) that involves driving on public roads to validate the system statistically. The motivation of this distance-based validation is to uncover rare events that are not captured during the validation process. Such occurrences are by definition rare events and are not uniformly distributed, meaning that the AV will need to drive a large number of miles\cite{mobileye_rss}\cite{elrofai2016scenario}\cite{kalra2016driving}. In addition to that, the collected evidence will be undermined with every software change \cite{mobileye_rss}\cite{elrofai2016scenario}\cite{kalra2016driving}.
Thus, this approach is considered to be time-consuming and expensive\cite{mobileye_rss}\cite{menzel2018scenarios}. 

Recently, the research community and automotive industry have broadly started to support scenario-based testing as an alternative strategy \cite{zhao2016accelerated}\cite{amersbach2019functional}\cite{de2017assessment}\cite{enable2016enable}\cite{putz2017system}. This has been considered as a feasible solution to realising HAVs on public roads. Most of the driving data collected in N-FOT is considered non-critical, thus not meaningful for validation purposes. The scenario-based approach for validating HAVs is promising as it focuses on critical scenarios without the need of driving millions of kilometres \cite{elrofai2016scenario}\cite{amersbach2019functional}. Scenario-based testing has been implemented in real-world and simulated settings. The detailed definition of the scenario and different classifications are explained in the methodology section. 
The aim of this paper is to find critical, concrete scenarios as defined by this classification. We use the term `scenario' to refer to a concrete scenario, and the terms `test case', `challenging scenario`, `challenging case`,`critical scenario' and `edge cases' are all used interchangeably.

From previous works \cite{zhao2016accelerated}\cite{de2017assessment}\cite{enable2016enable} in the area of scenario-based testing, the frequency distributions of parameters that are considered influential are extracted from N-FOT. From these, concrete scenarios are generated based on observation drawn from the distribution. The system gets evaluated by running a Monte-Carlo simulation using generated test cases and importance sampling using the test cases that were identified as critical in the Monte-Carlo run. 

In our work, we considered that scenarios are constructed using different parameters \cite{zhao2016accelerated}\cite{amersbach2019functional}\cite{de2017assessment} and that critical scenarios are considered as a combination of parameter values found to maximise the safety risks. The parameter optimisation technique is defined as the approach to find the the combination of parameter values that create challenging cases using an optimisation engine. The main contribution of this paper is to propose a innovative falsification method by considering the generation of challenging scenarios as a parameter optimisation problem. The falsification approach focuses on finding challenging cases in order to evaluate the System Under Test(SUT)\cite{tuncali2019rapidly}. The whole approach is considered as an optimisation guided falsification method, and the optimisation process looks to find the critical scenario that has the maximum return. The proposed method is based on Neural Architecture Search(NAS)\cite{nas}, which is the search for the best neural network architecture for a specific task. The NAS architecture is based on a policy gradient reinforcement learning (RL) algorithm. We have designed a reward function for the RL process in such a way that a positive reward for the combination of considered parameter values that can create challenging scenarios. In this experiment, we use the well established Responsibility-Sensitive Safety (RSS) metric\cite{mobileye_rss}\cite{rss}, the euclidean distance of pedestrian w.r.t ego-vehicle, and whether there was a collision as the metrics. The pedestrian crossing illustrated in figure~\ref{fig:ped_cross_sample} is used as the experimental context in the simulator to test our method.  

We can summarise the contribution of this papers as follows:

\begin{itemize}
  
  \item We propose an innovative falsification method to search for critical scenarios using a parameter optimisation technique based on NAS as the optimisation engine. The learning of the optimisation engine searches for critical scenarios that has maximize the measure of safety risk. Due to the RL strategy, the proposed method has better performance in searching for challenging scenarios compared to random search as used in many recent scenarios-based testing approaches\cite{zhao2016accelerated}\cite{de2017assessment}\cite{enable2016enable}\cite{tuncali2018simulation}, or brute-force search.
  
 \item We demonstrate that the proposed approach is scalable as it can handle high-dimension problems compared to our previous work\cite{karunakaran2020efficient}. Also, it can incorporate more parameters in a computationally efficient manner as opposed to the prior work.
  
 \item The proposed method can create more realistic scenarios compared to similar work in this area\cite{karunakaran2020efficient}\cite{koren2018adaptive}.
  
\end{itemize}

\section{Related work}

ISO 26262 is a widely accepted functional safety standard in the automotive industry, and the focus of this standard is on conventional vehicle systems\cite{iso26262}. This standard looks at preventing malfunctions(faults and failures) which are mainly related to internal causes\cite{kirovskii1}. However, the most functionally safe vehicle can still have an accident due to the performance limitations of the system caused by unintended behaviour when the system interacts with the external environment\cite{mobileye_rss}. The recently released ISO 21448 addresses this problem by aiming to reduce the number of unintended behaviours that a system can produce\cite{iso21448}. The standard suggests that safety of intended functionality is ensured by explicitly evaluating the known-unsafe scenario and evaluating unknown-unsafe scenarios using industry best practice, systematic analysis, or dedicated experiments\cite{iso21448}. The goal of the standard is to reduce the known unintended behaviours (known unsafe) and the unknown potential behaviour (unknown-unsafe) to an acceptable level of residual risk\cite{aptiv}.  These existing standards are essential to provide best practices for developing safety systems, but they are not comprehensive enough to cover the safety of Highly Automated Vehicles\cite{koopman2019safety}. These standards assume the ultimate guarantee of safety is the responsibility of the human driver. This is a fundamental problem for HAV operating on the road without a human driver. There are many technologies used in HAV that are inherently incompatible with existing automobile safety standards\cite{koopman2016challenges}\cite{aptiv}\cite{koopman2019safety}.

A common strategy is to use a test matrix for system validation\cite{zhao2016accelerated}. In this approach, the system is evaluated using pre-defined test scenarios extracted from the crash database\cite{zhao2016accelerated}. One such example is the AEB Autonomous Emergency Braking test protocol developed as part of the Euro New Car Assessment Program (Euro-NCAP)\cite{zhao2016accelerated}\cite{euro2013euro}.  Even though this approach is repeatable and reliable, these scenarios are fixed and predefined\cite{peng2012evaluation}. So, the system is likely to perform well in these tests, but the real-life scenarios are more diverse and challenging\cite{zhao2016accelerated}. 

Evaluating the performance of the system in a real-world environment is an essential aspect of robotic system design. Several Naturalistic Field Operation Tests(N-FOT) have been conducted around the world to evaluate the performance of automated vehicles using real-world data\cite{zhao2016accelerated}. In a real-world setting, exposure to rare events is uncommon\cite{mobileye_rss}\cite{zhao2016accelerated}. So this naturalistic data based approach or distance based approach needs to collect a large number of miles of data to make statistically significant observations\cite{mobileye_rss}. Although real-world testing is an important step in the development process, these approaches are time-consuming and expensive\cite{mobileye_rss}.

Recently, a significant body of research focused on scenario-based testing has been presented as alternative option to validate the SUT compared to naturalistic data based approach\cite{elrofai2016scenario}\cite{menzel2018scenarios}\cite{erdogan2018parametrized}\cite{weber2020simulation}. This type of approach is based on the assumption that most scenarios in the real world are non-critical and there is a need to generate critical scenarios intentionally\cite{weber2020simulation}\cite{amersbach2019defining}. Hence, it can reduce the required validation effort as a scenario based approach is focused directly on the relevant critical scenarios\cite{weber2020simulation}\cite{amersbach2019defining}. In \cite{de2017assessment}, scenarios are parameterised and stored in a database. Then, Monte-Carlo simulations are run to generate test cases from the parameterised scenarios. Similarly, in \cite{zhao2016accelerated}, frequency distributions are extracted from the real-world data, and Monte-Carlo simulations are used to generate test cases by sampling from these distributions. 
The ENABLE-S3 project\cite{enable2016enable} utilises the work from \cite{elrofai2018scenario} for the validation process. A method to extract the scenarios from real-world data is demonstrated by parameterising these scenarios. Then test cases can be derived from the frequency distribution generated during the parameterisation. Similarly, in Pegasus\cite{putz2017system}, data from different sources are used to cluster similar scenarios together to generate logical scenarios. Test cases (concrete scenarios) are created based on the logical scenarios. One of the challenges of scenario-based testing is parameter space explosion \cite{amersbach2019functional}. The scenario-based testing approach has to consider a high number of influential parameters. This leads to a high dimensional parameter space for validation, resulting in parameter space explosion. In other words, this approach requires a large number of concrete scenarios per logical scenario. In \cite{amersbach2019functional}, a functional decomposition method is proposed to reduce the effect of the parameter space explosion.

Waymo recreates the scenarios from the real-world data in simulation and generates thousand of variations of the recreated scenarios to validate the system\cite{waymo2017road}. They heavily rely on simulation to identify the critical events that are not encountered during real-world testing by creating variations of each scenario. This approach enables them to release software updates quickly rather than requiring millions of additional miles for validation.

\cite{koren2018adaptive} presents an interesting approach for validating the autonomous vehicles at a pedestrian crossing. They have proposed a method to find for most likely failure scenarios using a process called Adaptive Stress Testing. The primary strategy of the approach is to search for the most critical scenarios. Nevertheless, as there are no constraints on the variation of parameters this approach could generate non realistic scenarios, with non-smooth trajectories from the simulated participants. 

In our work, an efficient falsification approach to evaluate the SUT is proposed. The method searches for critical scenarios by considering the generation of scenarios as a parameter optimisation problem. The scenarios are constructed by varying the parameters and applying a loss function to optimise towards challenging scenarios. The approach is an optimisation-guided falsification method, and the optimisation process tends the scenarios towards the challenging cases with the maximum cost. In this work, we use NAS as an optimisation engine to create challenging scenarios. Figure~\ref{fig:overview_search} is the high-level overview of the proposed method where the search space represents the selected parameters and the search strategy is the optimisation approach. A new scenario is generated in the simulation using a mixture of the values for each parameter. The reward function is used to determine the level of safety risk in the generated scenario created from a new set of generated parameters. The reward function used is based on a combination of the Responsibility Sensitive Safety (RSS) metric\cite{mobileye_rss}, the euclidean distance of pedestrian w.r.t ego-vehicle, and the instance of a collision.

\section{Methodologies}

\subsection{Scenario definition and classification}

\begin{figure}[t]
\includegraphics[width=0.98\columnwidth]{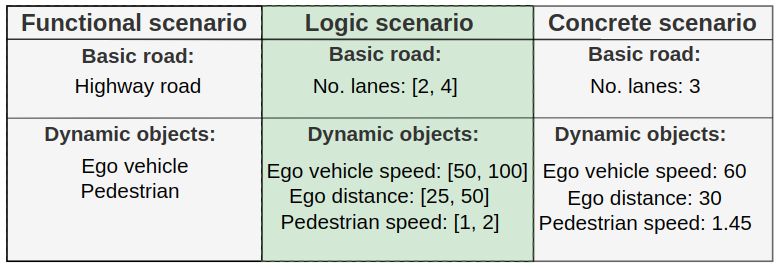}
\caption{\small Example of three main classification of scenarios. }
\label{fig:scenario_types}
\end{figure}

In \cite{elrofai2016scenario}, the scenario is defined as sequence of events. An event is defined as actions and manoeuvres performed by the host vehicle and other traffic participants within a scenario. Therefore, a scenario can consist of one or more events that are typically performed by the host vehicle (turn, lane
change, brake) and the other interacting traffic participants (cutting-in,
cutting-out, braking). In addition, a scenario can contain information about, for example, the
weather conditions (sunny, rainy), type of the road, state of the driver and the static
environment. A scenario is classified into three categories as shown in figure~\ref{fig:scenario_types}: functional scenario, logic scenario, and concrete scenario\cite{menzel2018scenarios}. The functional scenario is concerned about a high-level aspect of the scenario. For instance, types of road, list of traffic participants etc. are considered to be components of a functional scenario. We define a logic scenario as a functional scenario when a parameter range is specified.  This could be, for example, ego-vehicle speed between 50 kph to 100 kph and pedestrian speed between 1 to 2 meters per second. The concrete scenario is defined when we take concrete values for each parameter defined in the logic scenario. For instance, 60 kph for ego-vehicle speed and speed of 1.1 m/s for pedestrians. Finally, adding a pass or fail criteria to concrete scenarios is called a test case. 

\subsection{Markov Decision Processes and Reinforcement Learning}
A Markov Decision Process (MDP) is a Markov Reward Process (MRP) with decision capabilities. It represents an environment in which all of the states hold the Markov property, meaning that the future is independent of the past given the present\cite{sutton2018reinforcement}. The solution of an MDP is an optimal policy that evaluates the best action to choose from each state\cite{sutton2018reinforcement}. In MDP, at each time step $t$, the agent takes an action $a_t$ in the environment. Then, it receives the observation from the environment in the form of a new state $s_{t+1}$ given the current state $s_t$. State $s_t$ is the state representation of the agent constructed by either directly or indirectly observing the environment state\cite{sutton2018reinforcement}. The reward $r_t$ is a scalar quantity to give feedback to the agent about how appropriate (given the cost function) the action $a_t$ was based on the state $s_t$ from the environment. The policy is a mapping from states to actions\cite{sutton2018reinforcement}. In other words, policy determines how the agent behaves from a specific state. A value function or state-value is the expected total reward, starting from state $s$ and acting according to the current policy\cite{sutton2018reinforcement}. An action-value is the expected return for an agent starting from state $s$ and taking an arbitrary action before continuing on according to the current policy\cite{sutton2018reinforcement}. 

A Reinforcement Learning(RL) algorithm finds an optimal policy that maximizes the reward by interacting with the environment that is modelled as an MDP where no prior knowledge about the MDP is available\cite{sutton2018reinforcement}\cite{van2012reinforcement}. 

\subsection{REINFORCE}
REINFORCE implements Monte-Carlo sampling of a policy gradient method\cite{levine2017cs}. That means the RL agent samples from the starting state to generate the goal state directly from the environment, in comparison to other techniques that use bootstrapping such as Temporal Difference Learning and Dynamic programming. A policy gradient algorithm is a policy iteration approach where the policy is directly manipulated to reach the optimal policy that maximises the expected return. This type of algorithms is known as model-free RL. The model-free indicates that there is no prior knowledge of the model of the environment. In other words, we do not need to explicitly know the environment dynamics or transition probability. 
Pseudocode for the REINFORCE as follows\cite{levine2017cs}:
\begin{enumerate}
\item Sample N trajectories using policy $\pi_{\theta}$
\item Evaluate the gradient of the objective function J using the below expression:
\begin{equation} 
\label{eqn:reinforce1}
\nabla J(\theta) \approx \frac{1}{N}\sum_{\tau\in N}\sum_{t=0}^{T-1}\nabla_\theta log \pi_{\theta}(a_t, s_t) R(\tau)
\end{equation}
 Where N is the number of trajectories and R is the total return of a trajectory
\item Update the policy parameters
\begin{equation} 
\label{eqn:reinforce2}
\theta = \theta+\alpha \nabla J(\theta)
\end{equation}
\item Repeat 1 to 3 until we find the optimal policy $\pi_{\theta}$.

\end{enumerate}

The cross-entropy loss in the Neural Network(NN) can be used as a grad-log policy for the policy gradient expression. It opens up the use of NN as the functional approximator to represent the stochastic policy $\pi_{\theta}$.

\subsection{Neural Architecture Search (NAS)}

\begin{figure}[t]
\centering
\includegraphics[width=0.85\columnwidth]{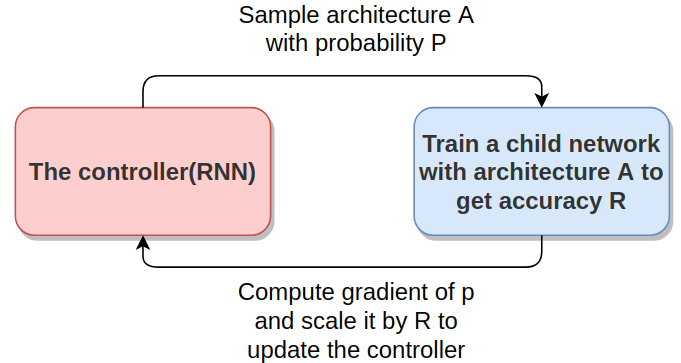}
\caption{\small Overview of Neural Architecture Search(NAS)\cite{nas}}
\label{fig:nas1}
\end{figure}

NAS is a hyperparameter optimisation approach to design a neural network for a specific task\cite{nas}.  As shown in figure ~\ref{fig:nas1}, the controller generates the architectural hyperparameters of the neural networks. The controller is implemented as a recurrent neural network. The whole problem is set up as a reinforcement learning problem to optimise the controller to generate the combination architectural hyperparameters to design the NN that maximises the reward. In \cite{levine2017cs}, a policy gradient RL method called REINFORCE is presented to solve the optimisation process.

\subsection{Responsibility Sensitive Safety (RSS)}

The ultimate goal of RSS as a formal method is to guarantee that an agent will not cause an accident, rather than to guarantee that an agent will not be involved in an accident\cite{mobileye_rss}. RSS has formalized the set of rules using four realms: Safe Distance, Dangerous Situation, Proper Response, and Responsibility\cite{rss}. A safe distance is calculated longitudinally and laterally based on the formula provided by the method. This method considers the worst-case scenario which eliminates the need for estimating road user intentions. In this paper, we focus on the longitudinal safe distance. By definition, the longitudinal safe distance is the minimum distance required for the ego vehicle to stop in time if a vehicle or object in front brakes abruptly.

\begin{equation} \label{eq1}
d_{min}=\left[v_r\rho+\frac{1}{2}a_{max,a}\rho^2+\frac{(v_r+\rho a_{max,a})^2}{2a_{min,b}}-\frac{v_f^2}{2a_{max,b}}\right]
\end{equation}

where $\ d_{min} $ represents the longitudinal safe distance, $\ v_r $  and $\ v_f $ are the velocity of the agent vehicle and front vehicle, respectively. $\ a_{min,b}$ is the minimum reasonable braking force of the agent vehicle, $\ a_{max ,b}$ is the maximum braking force of the front vehicle. In terms of acceleration, $\ a_{max,a} $ is the maximum acceleration of the front vehicle. $\rho$ is the agent vehicle response time.

Based on the safe distance calculation, RSS can determine whether or not the ego vehicle is in a dangerous situation. As per the RSS rule, the proper response should be evaluated and executed once it is determined that the vehicle is in a dangerous situation. If the ego-vehicle that follows the RSS involved in an accident, RSS can give an assurance that the responsibility of the accident does not belong to the ego-vehicle.

\subsection{Carla Simulator}

\begin{figure}[h]
\centering
\includegraphics[width=0.95\columnwidth]{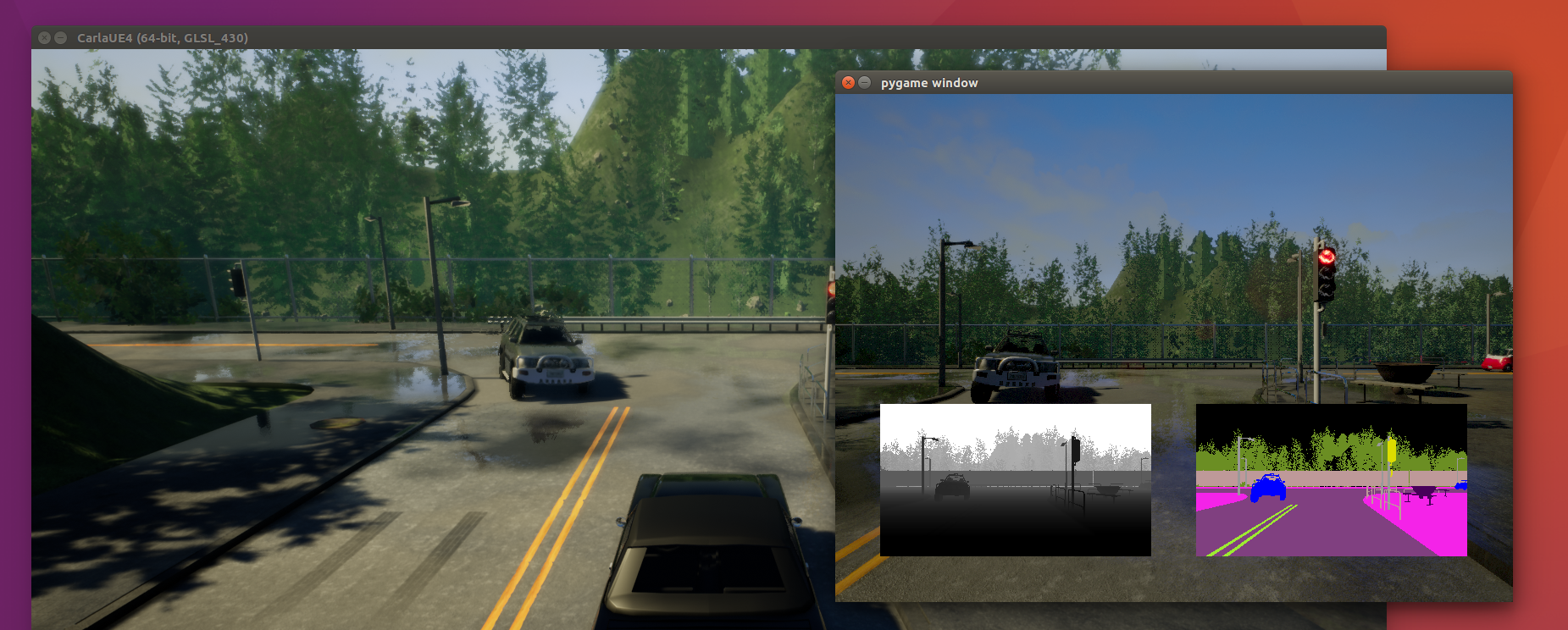}
\caption{\small Carla simulator}
\label{fig:carla_simulator}
\end{figure} 

Carla is an open source simulator for the development, training, and validation of autonomous urban driving systems\cite{carla_simulator}. The simulator consists of realistic participants and objects to represent the scenarios which are similar to the real world. The provided ROS bridge allows the extraction of data from the simulation directly in ROS format. As our real world research vehicle platform uses ROS, any algorithms from the real vehicle can be connected through the ROS bridge to operate in Carla. This allows us to test any sub-system in simulation that is used in the real vehicle. The ROS bridge has been modified for the purpose of building the proposed method.

\subsection{Proposed method}

\begin{figure}[t]
\centering
\includegraphics[width=0.85\columnwidth]{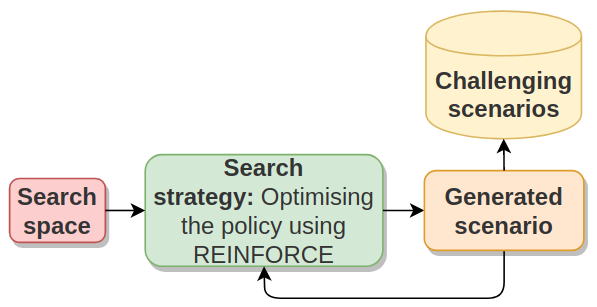}
\caption{\small This figure depicts a high-level overview of the proposed falsification method. The search space represents the range of possible parameter values. The search strategy for the optimisation process is a policy gradient RL called REINFORCE to generate scenario in each episode and to learn towards the challenging case that has maximum return.}
\label{fig:overview_search}
\end{figure}

The proposed method is a falsification method based on a parameter optimisation technique to search for the challenging scenarios of the SUT. As shown in Figure~\ref{fig:overview_search}, the parameter values are defined in the search space. Our search strategy is to optimise the search to find the combination of parameter values that leads to challenging cases. This is done using a policy-gradient reinforcement learning algorithm. The whole approach comes under optimisation guided falsification, and the learning process is towards the challenging case that has the maximum return.

\begin{figure}[t]
\includegraphics[width=0.98\columnwidth]{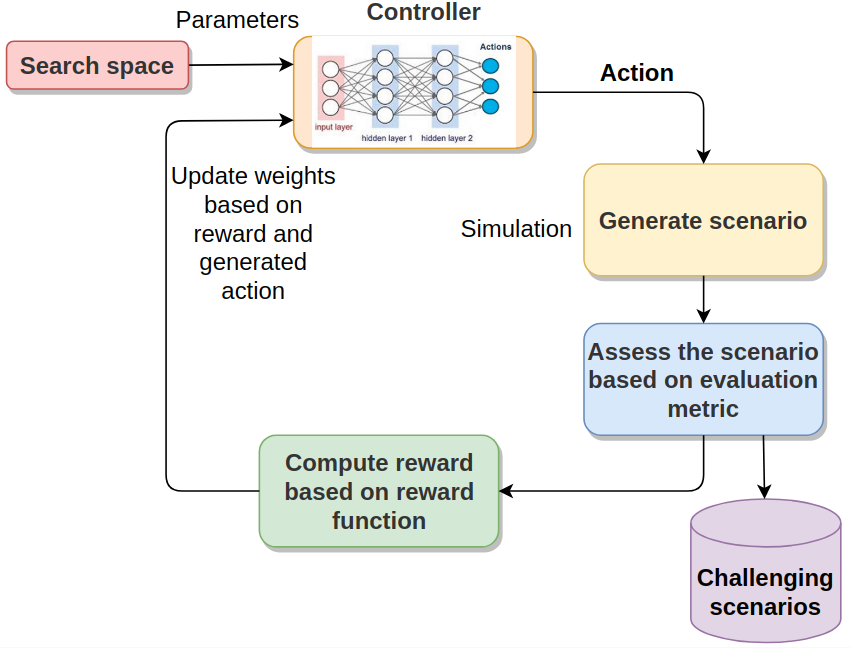}
\caption{\small This figure illustrates the detailed architecture of the proposed method—the overall process considered as Policy gradient RL. The purpose is to generate challenging scenarios by optimising the controller to produce actions that leads to a cases with the maximum return, indicating scenarios with higher safety risk. The reward is used to update the weights of the controller. Each action is a combination of parameter values that are taken from the search space.}
\label{fig:architecture}
\end{figure}

As shown in figure~\ref{fig:architecture}, the architecture of the proposed method is a modified version of the original Neural Architecture Search(NAS) \cite{nas}. The overall idea is to generate  critical scenarios by optimising the controller using the policy gradient algorithm. 
The action produced by the controller in each episode is the combination of the parameter values. 
The end result is that the controller will optimise the parameters to tend towards the scenarios with the highest risk by maximising the cost function.
During the initial stages, the controller has not yet learned suitable parameter combinations, so the actions initiated by the controller may not create challenging cases. At later stages, the controller has already been optimised to produce the right mixture of values as an action. In each episode, the controller outputs the actions as a set of parameter values. Based on these values, a scenario is constructed in the Carla simulator. The simulation is carried out to compute the reward, determining the level of safety risk generated by the action. After the episode ends, the action and reward is stored.  The controller is then updated in the direction of maximum rewards after N episodes. More specifically, we are using a Recurrent Neural Network(RNN) as the controller, so the weight of NN gets updated based on the gradient of the objective function.

The proposed method is an efficient falsification approach to generate critical scenarios for evaluating the SUT as compared to many random-search based scenario-based testing and brute-force testing to finding these scenarios. In this approach, we consider the value range for each parameter instead of a constant value. The optimisation process is over all of these parameters. As the proposed method handles this relatively high-dimensional problem, it is more scalable than our previous work. In addition, the method is computationally efficient enabling it to incorporate more parameters. The proposed system can create more realistic scenarios by considering the existing parameter frequency distribution from previous works to define the parameter range; for instance, pedestrian velocity and acceleration values can be obtained from frequency distributions described in earlier works\cite{chandra2013speed}\cite{liu2017microscopic}. Incorporating these constraints allows for the creation of smooth trajectories, which results in more realistic scenarios.

\section{Experiment}

\begin{figure}[h]
\centering
\includegraphics[width=0.95\columnwidth]{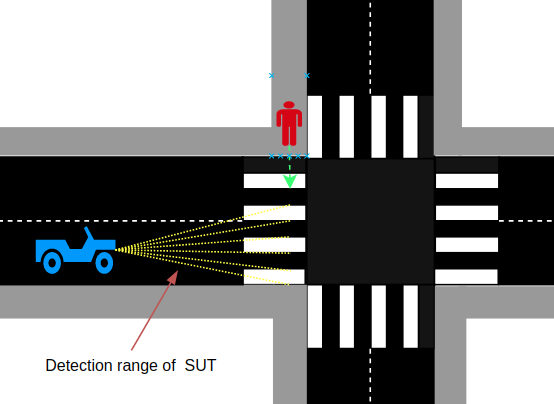}
\caption{\small Pedestrian crossing intersection}
\label{fig:pedestrian_crossing}
\end{figure} 

In this work, we conducted experiments at a pedestrian crossing using the Carla simulator, as shown in Figure~\ref{fig:pedestrian_crossing}. We have implemented the proposed method, as depicted in Figure~\ref{fig:architecture}. The goal of this method is to evaluate the SUT using a falsification approach by searching for challenging cases using a parameter optimisation technique. The overall process is to teach the controller to predict the suitable action that can create critical scenarios by biasing towards the scenario that has the maximum return. At each episode, the controller predicts an action that is a combination of parameter values from search space. Based on the list of parameters in the action, a concrete scenario or test case is constructed. Once the scenario is completed, a reward function will be used to validate the quality of the action from the controller. The reward function is then used to update the controller.

We have selected a relatively simple task in order to demonstrate the method, with more complexity to be added in future work. We expect that this approach can be used to determine the critical combinations of parameter values for 
autonomous vehicle sub-systems such as collision avoidance systems, to create challenging cases when working under a particular ODD. At this stage, we are not considering the pedestrian intention and uncertainties from the ego vehicle's sensor measurement.

The main components of the experiment are explained below, before discussing the implementation of the proposed method.

\begin{figure}[h]
\includegraphics[width=0.95\columnwidth]{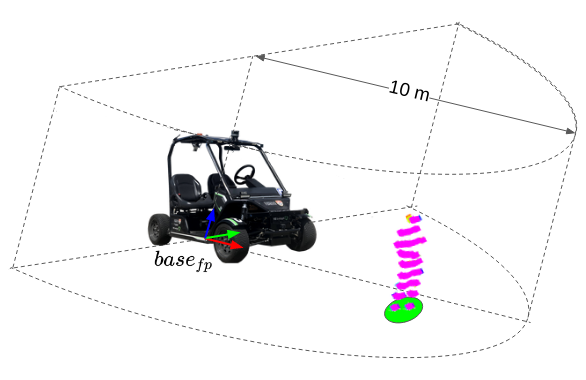}
\caption{\small Collision Avoidance System (CAS)'s detection range }
\label{fig:sut}
\end{figure}

\textbf{System Under Test (SUT)}: We are using a Collision Avoidance System (CAS) provided by the Carla simulator to test our method. As depicted in the Figure~\ref{fig:sut}, the maximum detection range of the system is 10 meters. The systems basic functionality is to control the speed of the vehicle to avoid a collision when an object is detected. Importantly, we are treating this system as a blackbox system.  

\textbf{Search space}: The search space contains the parameters and their frequency distribution to create the scenarios. We have considered 5 parameters in this experiment: Ego-vehicle longitudinal start position (ego-long-pos), pedestrian acceleration (ped-accel), pedestrian velocity (ped-vel), pedestrian longitudinal start position (ped-long-pos), and weather. Carla has 15 weather settings, and values from the weather parameter indicates each of these settings. The list of parameters and their distribution is shown in the Table~\ref{table:search_space}. No. of samples is an indicator of how many values are sampled from the distribution for each parameter.

\begin{table}[h!]
\centering
 \caption{List of parameter in the search space}
 \label{table:search_space}
 \begin{tabular}{|c c c|} 
 \hline
 Parameter & Distribution & No. of samples \\ [0.5ex] 
 \hline\hline
 ego-long-pos & $\mathcal{U}(1,10) m$ & 10\\ 
 ped-accel & $\mathcal{U}(0,0.1) m/s^2$ & 10\\
 ped-vel  & $\mathcal{N}(1.46,\,0.24) m/s$ & 25\\
 ped-long-pos & $\mathcal{U}(3,4.5) m$ & 4\\
 weather & $\mathcal{U}(0,14)$ & 10\\ [1ex] 
 \hline
 \end{tabular}
\end{table}

The samples taken from the distributions in this experiment as follows:

ego-long-pos:$[1, 2, 3, 4, 5, 6, 7, 8, 9, 10]\\$

ped-accel: $[0.046, 0.051, 0.097, 0.075, 0.099, 0.076, 0.065,\\ 0.007, 0.007, 0.013]\\$

ped-vel: $[1.803, 1.178, 1.139, 1.476, 1.205, 1.725, 1.142,\\ 1.516, 1.201, 1.614, 1.808, 1.303, 1.565, 1.416, 1.247, 1.355,\\ 1.755, 1.237, 1.196, 1.303, 1.552, 1.344, 0.937, 1.108, 0.976]\\$

ped-long-pos:$[3, 3.5, 4, 4.5]\\$

weather: $[4, 1, 7, 8, 6, 5, 8, 12, 9, 2]\\$

\textbf{State space and Action space}: The method is based on NAS\cite{nas} and both state space and action space represents the same item. The controller accept state $s_t$ and reward based on the appropriateness of the action $a_{t-1}$ and generates the next action $a_t$. State represents the previous action $a_{t-1}$.

\begin{multline}
Parameters,\ P = (ego-long-pos,\ ped-accel,\\ ped-vel,\ ped-long-pos,\ weather)
\end{multline}

\begin{equation}
a_{t} = P,\ s_{t} = a_{t-1}
\end{equation}

\textbf{Reward function}: This function determines the safety risk for an action given the state. At each time steps$\ t $, the minimum safe distance is measured using the RSS formula. This metric compares with the Euclidean distance to the pedestrian. The expected behaviour of the SUT is to reduce the speed of the ego vehicle when an object is detected. If the Euclidean distance is less than the safe distance measured by RSS, it is considered as an improper response for the SUT. We classify this as a high-risk timestep, $highrisk_{ts}^{RSS}$. At the end of the episode, a total number of high-risk timesteps get normalised between 0.01 and -0.01 using the equation~\ref{eqn:normalize2}. It is essential to normalise the reward to a small value range as a large range can affect the ability of the controller to learn.

\begin{equation}
\label{eqn:normalize2}
new_x = (b-a)*\frac{x-min}{max-min}+a
\end{equation}

Where $x$ is the normalised total number of improper responses by the SUT, minimum (min) value for the no. of high-risk timesteps and maximum(max) value is the total no. of timesteps in an episode. The new minimum value $a$ is -0.01 and the new maximum value $b$ is 0.01

A higher number of improper response by the SUT gives a high reward as the objective is to optimise the parameters towards the challenging cases. If there is a low number of high risk responses, this indicates that the system is less likely to have safety risks. 
If the combination of parameters lead to a collision, we reward an extra 0.25 to the total reward.

\begin{equation}
\label{eqn:normalize1}
new_x = \frac{x-min}{max-min}
\end{equation}

The third type of reward is based on the euclidean distance between pedestrian and ego-vehicle. When the distance gets smaller, it is more likely to be in a high risk situation due to RSS, or the potential of collision. At the end of the episode, the final distance is calculated and normalised between 0 and 1 using the equation~\ref{eqn:normalize1}. To provide a higher reward for a smaller distance, we reverse the normalised distance by subtracting the normalised distance from 1. Then, using the equation~\ref{eqn:normalize2}, we further normalise the distance between -0.01 and 0.01 for the reward. The final reward gets computed by adding the output of these three reward functions.

\begin{equation}
R = \Bigg\{\begin{matrix}-0.01\ to\ 0.01&\ highrisk_{ts}^{RSS}\\
-0.01\ to\ 0.01&\ euclidean\ distance\\
0.25&\ collision\\
\end{matrix}
\end{equation}

\textbf{Episode}: Typically, the episode is a sequence of states and actions. In this architecture, we have only one state and action per episode. A scenario is constructed from the parameter values in each action. The end of a scenario is determined either by time elapsed, distance travelled by ego vehicle, or in the event of a collision. Once the scenario ends, it is considered the end of the episode and the reward is computed. Each scenario is set to run up to a maximum of 4000 episodes. We set this number arbitrarily to demonstrate the proposed method.

\textbf{Evaluation metric}: The evaluation metric determines whether a scenario can be considered as challenging or non-challenging. The$\ distance_{Euclidean}$ is the euclidean distance between pedestrian and ego vehicle, and $\ distance_{RSS}$ is the minimum safe distance measured by RSS. We calculate both values at each timestep of an episode. If  $\ distance_{Euclidean}$ < $\ distance_{RSS}$, then it is marked as high-risk timestep. The high-risk timestep is represented as $\ highrisk_{ts}^{RSS}$. We consider the overall scenario as a critical scenario if the number of high-risk timesteps is greater than 50$\%$ of the total timesteps. Otherwise, it will be designated as a non-challenging scenario. We arbitrarily chose 50$\%$ as a threshold limit to demonstrate the proposed method, further study into the appropriate values for this number are part of future work. 
Another criterion to determine the challenging scenario is based on the collision. Even if the total $highrisk_{ts}^{RSS}$ is less than the threshold, a scenario is set to be a challenging scenario if there is a collision.

\begin{equation}
Challenging_{scenario} = \Bigg\{\begin{matrix}\frac{1}{N}\sum_{i=1}^{N}highrisk_{ts}^{RSS}*100\ >= \\50\% \\
or\\
Collision\\
\end{matrix}
\end{equation}

\textbf{Controller and training}: The learning aspect of the controller as shown in the figure~\ref{fig:architecture} governs by the policy gradient RL algorithm called REINFORCE. The goal of the algorithm is to optimise the controller by generating the action that leads to the most critical events. In RL terms, the controller learns the policy that maximises the return and the policy is represented in this experiment by the controller. In REINFORCE, a Neural Network(NN) can be used as a Universal Function Approximator to represent the policy. We use a Recurrent Neural Network(RNN) as functional approximator to represent the policy in this experiment. 

\begin{figure}[h]
\includegraphics[width=0.95\columnwidth]{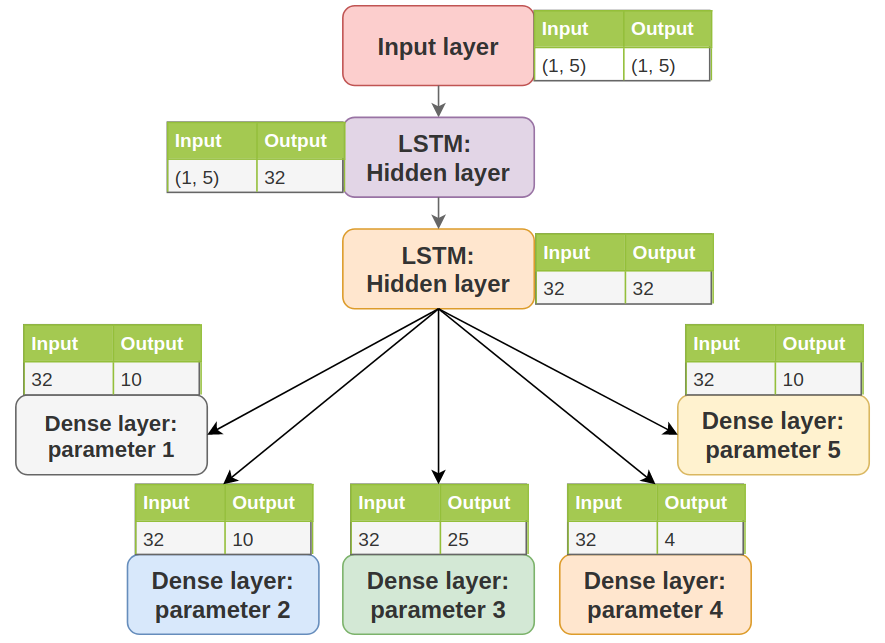}
\caption{\small Architecture of the controller}
\label{fig:controller_architecture}
\end{figure} 


As shown in the figure~\ref{fig:controller_architecture}, there is a dense layer for each output. In this experiment, five dense layers represent the five parameters. The number of output units in each dense layer varies based on the number of samples for each parameter. Each dense layer has a separate Categorical Cross-entropy loss function. We choose the Adam optimiser to train the network. 

The training process of the NN in REINFORCE is slightly different compared to a standard NN. Usually, in every epoch the gradient of the loss function is computed. Then weights get updated according to the direction of the gradient. In a supervised learning setting, we have the ground truth, and the loss function can compute by how much the prediction has deviated. In RL, there is no ground truth, and a reward is an indicator of how well the action generated the desired response. In this case, it is essential to update the weights in the direction of the reward. 

Instead of updating weights in every epoch, the RL approach waits for N episodes. Once N trajectory samples are collected (a trajectory contains state, action, and reward), the gradient of the loss functions is computed based on the equation~\ref{eqn:reinforce1} of the REINFORCE algorithm. The value of N is set to 25 episodes in this work. In equation~\ref{eqn:reinforce1}, the reward is multiplied with gradient log of the NN output. The NN represents the policy and the gradient log of the output is the gradient of the standard cross-entropy loss function. Based on the equation~\ref{eqn:reinforce2}, we can update the weights of the NN, $\theta$ with the gradient expression computed in the equation~\ref{eqn:reinforce1}.

In RL, the balance between exploitation vs exploration is an important aspect. 
Exploration at the early stages enables the model not to get stuck in a local optima.
Initially, an agent is weighted towards exploration compared to exploitation as determined by the $\ \epsilon$ value. The initial value of $\ \epsilon$ is set to 1, decaying at each time step by multiplying by 0.995 until $\ \epsilon_{min}$=0.01. During this time, the actions selected for the agent will be random. 

\textbf{Implementation}: The proposed method evaluates the SUT by searching for the challenging scenarios where the system fails to satisfy the safety requirements, or the evaluation metric threshold. The search process is completed by optimising the controller towards the critical case that has maximum return. Firstly, the search space is created by sampling from the parameter frequency distribution as shown in the table~\ref{table:search_space}. Once the search space has been defined, the iterative process of learning begins. 

At each episode, the controller generates the action given a state. During the exploration stage, the controller has not yet learned anything, so the output is towards a randomly generated action. Later, the controller learns the optimal policy to create an action that leads to a challenging scenario for the SUT. 

The generated action contains five parameter values.  The method uses these values to constructs a scenario in the Carla simulator. In this work, we are focusing on developing critical cases for the SUT in a pedestrian crossing. The realistic scenarios get constructed as the velocity and acceleration are sampled from the frequency distributions based on previous works. 
More realistic scenarios are constructed by incorporating the acceleration and velocity constraints which limits the change in pedestrian speed over sequences of timesteps. 

The reward is based on three criteria: RSS failure, euclidean distance, and collision. At the end of each episode, the final reward is computed by adding the individual rewards from these three criteria. A high reward is applied to the action that leads to failure (high risk) and low reward for the successful (low risk) scenarios.

As the state is the previous action, we set the last action as the current state for the controller. The simulator gets reset to generate the next scenario based on the following action. Action and reward are then stored for 25 episodes. The controller represents the policy, and after every 25 episodes, the policy gets updated using the REINFORCE algorithm to updating the network weights,$\theta$.

In this work, we have set the controller to operate for 4000 episodes to learn the optimal policy. Depending on the number of parameters and the values, we may need to adjust this limit. 

\section{Results}
\begin{figure}[h]
\includegraphics[width=0.9\columnwidth, trim={0.5cm 0 1.7cm 0},clip]{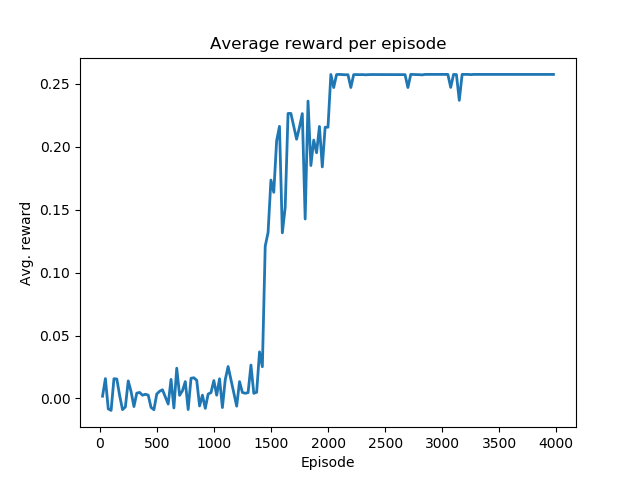}
\caption{\small Metric used to measure the performance of the method. The plot shows that after the 1500th episode, the approach learned the action that can obtain a maximum return. It indicates that the controller has started to predict the right action that maximises the return just over 1500th episode. From 2000th episode onwards, the reward is almost constant.}
\label{fig:reward_per_episode}
\end{figure}

The results of our approach are demonstrated using several metrics that show how the potentially unsafe challenging cases can be learnt using our method. The goal of this method is to evaluate SUT by searching for challenging scenarios in which the SUT fails to satisfy the evaluation metrics. 
As shown Figure~\ref{fig:reward_per_episode}, the proposed method begins to converge after the 1500th episode. The average reward remains almost constant from 2000th episodes onwards. It means that the controller has learned to predict the correct action that creates a challenging case with maximised return. In the brute-force search or enumeration technique, we need to go through $12^5 = 248832$ concrete scenarios where 12 is the average number of values in a parameter, and 5 is the total number of parameters. Similarly, in a random search based method, we do not know how many concrete scenarios needed to be considered. This result indicates that our method can evaluate the SUT by generating the challenging scenarios efficiently compared to the random search based scenario-based testing and brute force search.  

\begin{figure}[ht]
    \begin{subfigure}{\columnwidth}
      \centering
      \includegraphics[width=0.95\columnwidth, trim={0 0 0 0},clip]{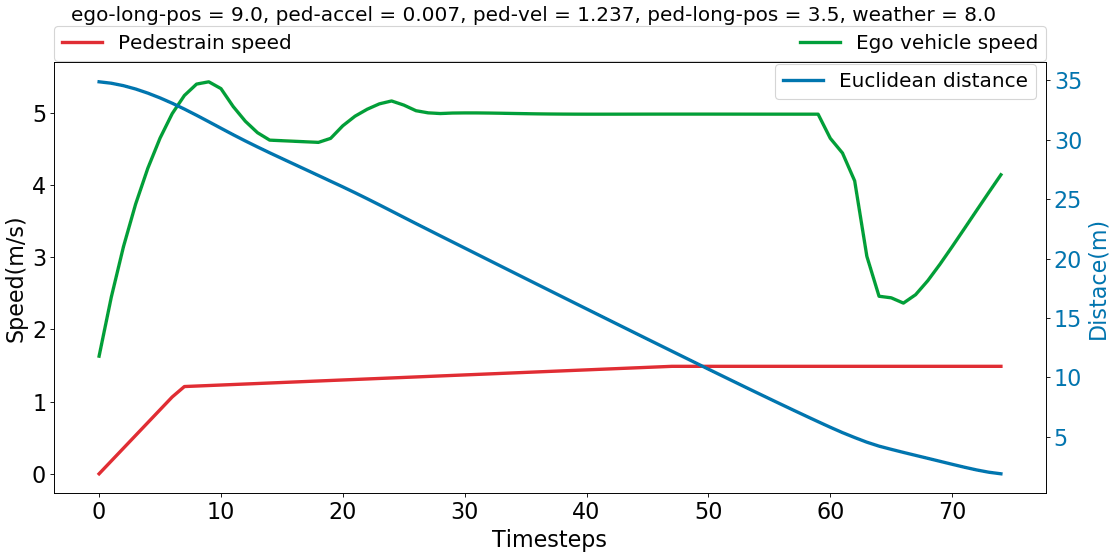}  
      \caption{\small This diagram depicts the challenging scenario that has a maximum return—this particular scenario ended up in a collision. The figure shows the speed of the pedestrian and ego-vehicle, and the parameter values that create the challenging case. The figure illustrates that ego-vehicle reduced the speed after seeing the pedestrian, but accelerated again which ended up causing a collision.}
      \label{fig:challenging_scenario}
    \end{subfigure}

    \begin{subfigure}{\columnwidth}
      \centering
      \includegraphics[width=0.95\columnwidth, trim={0 0 0 0 },clip]{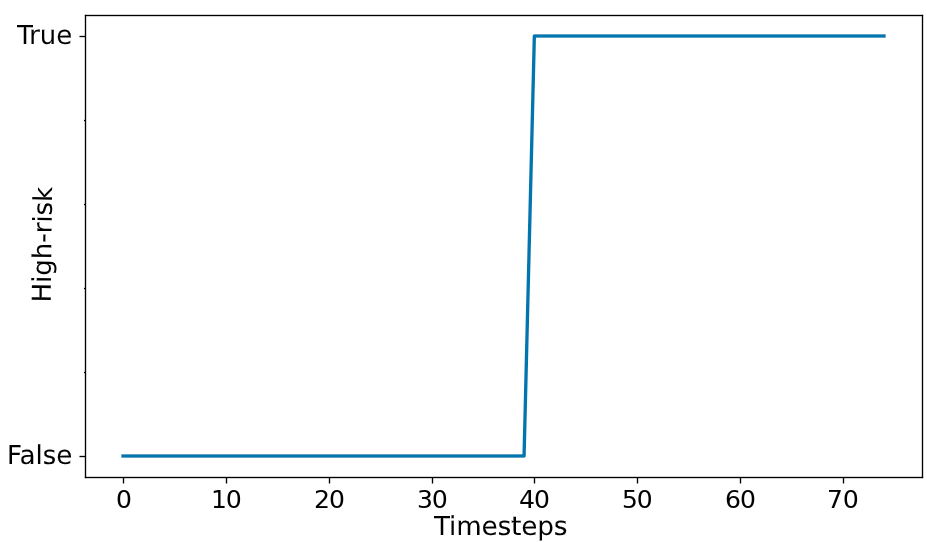} 
      \caption{\small This figure depicts the number of high-risk timesteps in a challenging scenario. It indicates that continuous high-risk timesteps can lead to collision or challenging scenarios based on the evaluation metric.}
      \label{fig:challenging_scenario_timesteps}
    \end{subfigure}
\caption{\small This plot illustrates the challenging scenario that finishes in a crash.}
\label{fig:fig_challenging_scenario}
\end{figure}

The definition of a challenging case is based on the evaluation metric. A typical challenging scenario that leads to the collision is depicted in figure~\ref{fig:challenging_scenario}. The diagram illustrates the pedestrian speed(red) and ego vehicle speed(green) at each time step of the scenario. Also, all of the parameter values that are required to create the scenario are shown at the top of the figure. This particular scenario is constructed using the action from the controller. The controller converged to the optimal policy towards the end of the learning process. 
In this scenario, collision occurred as a result of  the speed of the pedestrian being around 1.237 m/s, and acceleration of 0.007$m/s^2$ with the starting position of the ego vehicle at 9m and pedestrian starting position at 3.5m.
The contributing factor for this scenario is that the pedestrian managed to reach the centre of the crosswalk while ego-vehicle got closer to the pedestrian without reducing the speed. After reducing the speed, the ego-vehicle accelerated again causing the collision with the pedestrian. 

\begin{figure}[h]
      \includegraphics[width=0.95\columnwidth]{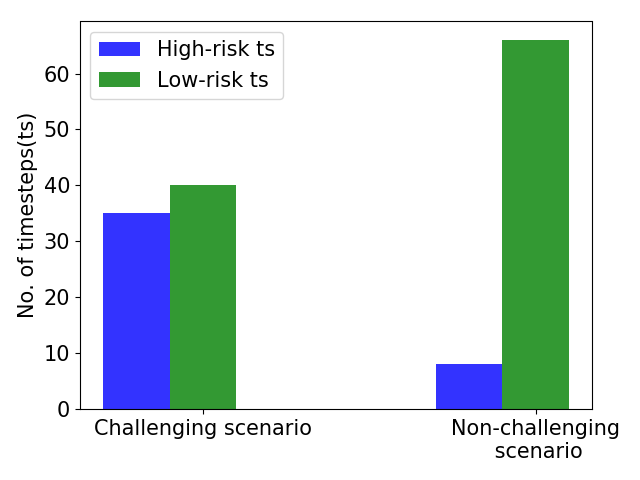} 
      \caption{\small The figure depicts the comparison between a challenging scenario and non-challenging scenario based on high-risk and low-risk timesteps. It is evident that the challenging scenario has more high-risk timesteps than the non-challenging scenario.}
      \label{fig:bar_chart}
\end{figure} 

We observed a collision after a number of continuous highrisk timesteps. Based on ~\ref{fig:challenging_scenario_timesteps}, continuous high-risk  timesteps occurred from the 39th timestep until the end of the episode. As shown in the figure~\ref{fig:bar_chart},  we can see the difference between the number of high-risk timesteps and low-risk timesteps for a challenging scenario and non-challenging scenario.

\begin{figure}[ht]
\vspace{3mm}
\begin{subfigure}{\columnwidth}
  \centering
  \includegraphics[width=\columnwidth, trim={0 0 0 0},clip]{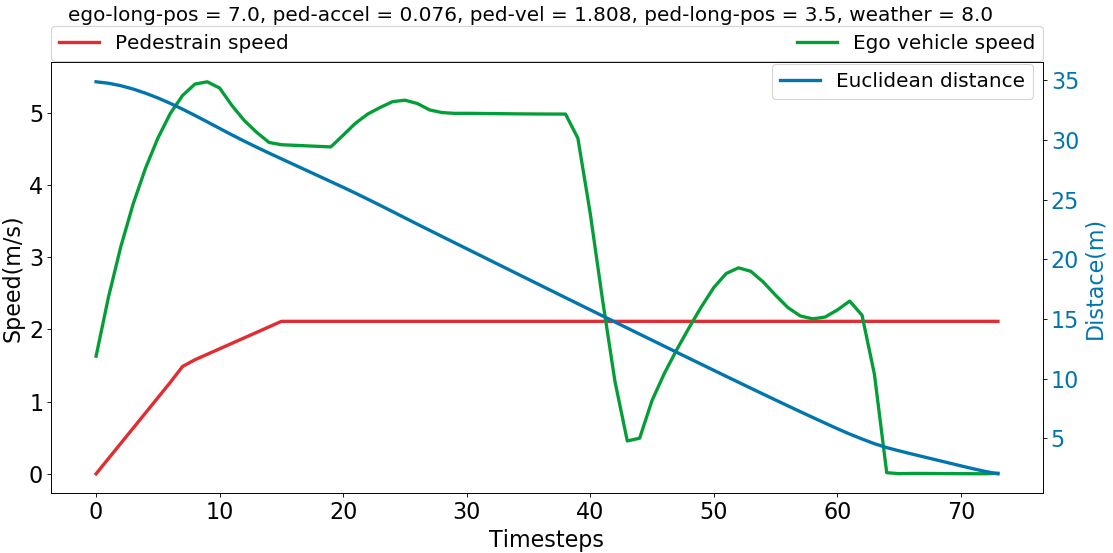}  
  \caption{\small The figure explains the typical non-challenging scenario in the experiment. Pedestrian reached at the centre of the crossing long before the ego-vehicle reached the crossing.}
  \label{fig:success_scenario}
\end{subfigure}
\begin{subfigure}{\columnwidth}
  \centering
  \includegraphics[width=0.95\columnwidth, trim={0 0 0 0},clip]{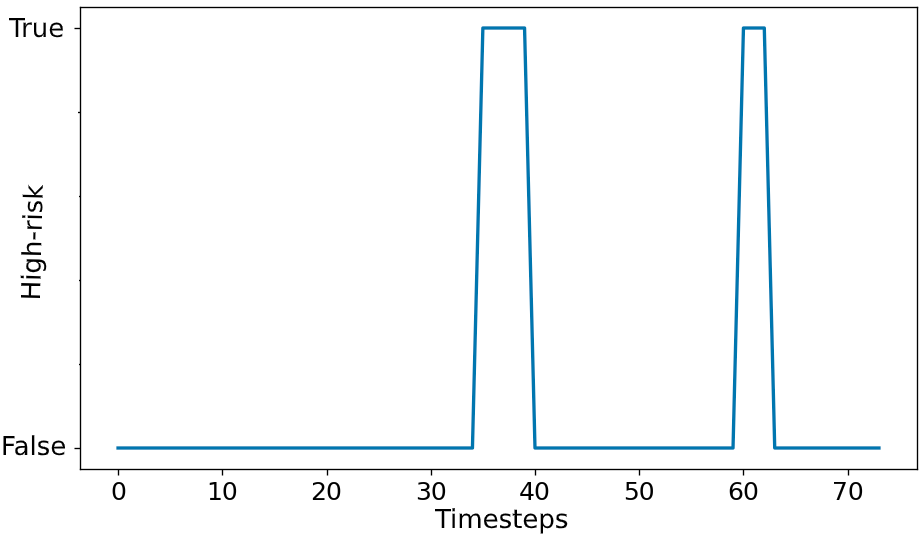}
  \caption{\small The diagram shows that there is a continuous high-risk timesteps for a few timesteps, but it's not consistent.}
  \label{fig:success_scenario_timesteps}
\end{subfigure}
\caption{\small This figure depicts the example of typical non-challenging scenario based on the evaluation metric.}
\label{fig:fig_scenario2}
\end{figure}

Figure~\ref{fig:fig_scenario2} illustrates the typical example of non-challenging scenario based on the evaluation metric.
In this case, a pedestrian walked almost at the same speed as in the challenging scenario with comparatively less acceleration. The contributing factor is that the acceleration and speed of the pedestrian are higher than the challenging scenario. As a result, the pedestrian reaches the centre long before the vehicle arrives in the crossing area. There is a smaller number of high-risk time steps and no collision, which defines this as a successful (i.e. low risk) scenario.

\begin{figure}[t]
\includegraphics[width=0.98\columnwidth]{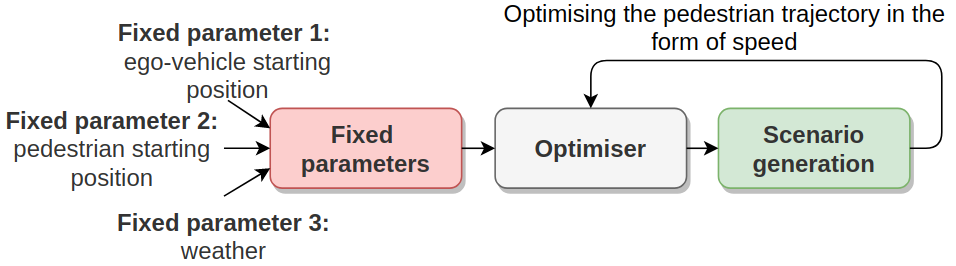}
\caption{\small The figure depicts the approach taken in our previous work\cite{karunakaran2020efficient}. The changes in the fixed parameters can lead to new different scenarios. As it considered only one parameter for the optimisation, the whole approach is solving the low-dimensional problem. This approach is not scalable with more parameters.}
\label{fig:scalable}
\end{figure}

The proposed method is a scalable approach compared to previous work\cite{karunakaran2020efficient}. In \cite{karunakaran2020efficient}, pedestrian speed was the only parameter that was varied for the optimisation process. The other parameters such as the starting position of ego vehicle, the pedestrian starting position, and weather were held constant. The issue here is that we do not know the optimal values for these constant parameters to generate challenging scenarios. 
This low dimensional representation of the problem limited the ability to find the challenging scenarios.
In our proposed method, the architecture enables us to add a set of parameters with distributions over realistic values. During the optimisation process, the proposed method searches over the parameter values. In this experiment, five parameters each defined for a valid range of values are considered. This approach is more scalable as it can handle the high-dimensional problem without exponential increase in computational requirements. 

\begin{figure}[h]
\includegraphics[width=0.9\columnwidth, trim={0.5cm 0 1.7cm 0},clip]{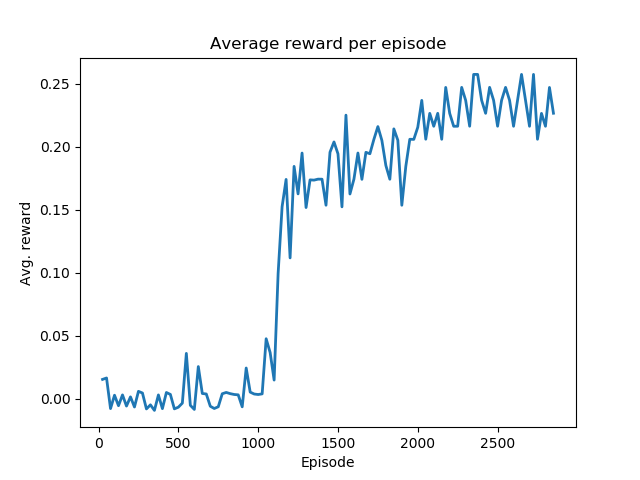}
\caption{\small The figure depicts the average reward per episode for the experiment to evaluate the scalability. We have added two more parameters to the search space to see whether the learning time has a significant change. It is evident from the figure that achieving the highest reward did not take a substantial number of episodes.}
\label{fig:reward_per_episode_scalable_new}
\end{figure}
To demonstrate the ability of the system to handle higher dimensional problems, we included two additional parameters which impact on the generation of the scenarios. The first additional parameter is designed to cause abrupt changes to the pedestrian speed for five timesteps. The second parameter was included to indicate a timestep in which the unexpected change was to commence. The values for two additional parameters added as follows:

ped_speed_change:$[0.50, -0.50, 0.75, -0.75]\\$

ped_timesteps: $[20, 30, 40, 50, 60]\\\\$

As shown in Figure~\ref{fig:reward_per_episode_scalable_new}, adding more parameters does not increase the number of episodes required to converge to an optimal policy that has maximum return. The average reward per episode is similar to the results in Figure~\ref{fig:reward_per_episode}, which was the experiment using five parameters. There is a substantial increases in the possible combination of parameter values with 5 parameters from  $12^5 = 248832$ to $9^7 = 4782969$ with 7 parameters.This demonstrates that our approach is scalable as the increase in parameters does not increase the computational requirements by a significant amount.

One of the contributions of the proposed method is to create more realistic scenarios by using the parameter distributions measured from existing real-world data.
As shown in the figures~\ref{fig:fig_challenging_scenario} and~\ref{fig:fig_scenario2}, the speed and acceleration of the pedestrian is more realistic as compared to the previous work \cite{karunakaran2020efficient}\cite{koren2018adaptive}. The acceleration and velocity is bounded for the pedestrian motion while they are crossing the road, better reflecting the real world physical constraints. The pedestrian trajectory is created at every episode based on the selected value of the velocity parameter, ped-vel and acceleration parameter, ped-accel. As shown in Figures~\ref{fig:challenging_scenario} and~\ref{fig:success_scenario}, the pedestrian trajectories are very similar, the main difference being the differences in speed and acceleration. 
Using trajectories based on real world physical constraints makes the validation of SUT more efficient as iterations of the optimiser are not wasted on unrealistic and infeasible trajectories.

\section{Discussion and Conclusion}

This paper presented a falsification method to evaluate a system under test with a number of influencing parameters that is more efficient when compared to existing random search or brute-force search based methods. 
Overall, the scenarios generated by the proposed method are more realistic than previous works as appropriate distributions for the parameters representing the pedestrian kinematics are incorporated to produce smooth, feasible pedestrian trajectories.

In future work, we will be increasing the complexity of the experiments by incorporating more parameters to evaluate the scalability of the algorithms further. Using our experimental autonomous vehicle platforms, we hope to assess the on-board safety systems to compare the performance to the simulated results found in this paper. Finding critical scenarios with a specific design indicates that we have to redesign the SUT to provide a higher standard of safety. We aim to demonstrate that we can effectively influence the design of SUT using this approach.


\section*{Acknowledgment}
This  work  has  been  funded  by  the  Australian  Centre  for Field Robotics (ACFR), University of Sydney and Insurance Australia Group (IAG) and iMOVE CRC and supported by the  Cooperative  Research  Centres  program,  an  Australian Government initiative.

\bibliographystyle{IEEEtran}
\bibliography{references.bib}

\begin{thebibliography}{10}
\providecommand{\url}[1]{#1}
\csname url@samestyle\endcsname
\providecommand{\newblock}{\relax}
\providecommand{\bibinfo}[2]{#2}
\providecommand{\BIBentrySTDinterwordspacing}{\spaceskip=0pt\relax}
\providecommand{\BIBentryALTinterwordstretchfactor}{4}
\providecommand{\BIBentryALTinterwordspacing}{\spaceskip=\fontdimen2\font plus
\BIBentryALTinterwordstretchfactor\fontdimen3\font minus
  \fontdimen4\font\relax}
\providecommand{\BIBforeignlanguage}[2]{{%
\expandafter\ifx\csname l@#1\endcsname\relax
\typeout{** WARNING: IEEEtran.bst: No hyphenation pattern has been}%
\typeout{** loaded for the language `#1'. Using the pattern for}%
\typeout{** the default language instead.}%
\else
\language=\csname l@#1\endcsname
\fi
#2}}
\providecommand{\BIBdecl}{\relax}
\BIBdecl

\bibitem{ITS_2020_workshop}
M.~Soledad~Elli, I.~Alvarez, T.~Stolte, F.~Wirth, C.~Stiller, M.~Maurer, and
  J.~Weast, ``Workshop on automated vehicle safety: Verification, validation
  and transparency,'' in \emph{2020 IEEE Intelligent Transportation Systems
  Conference (ITSC)}.\hskip 1em plus 0.5em minus 0.4em\relax IEEE, 2020.

\bibitem{mobileye_rss}
\BIBentryALTinterwordspacing
S.~Shalev{-}Shwartz, S.~Shammah, and A.~Shashua, ``On a formal model of safe
  and scalable self-driving cars,'' \emph{CoRR}, vol. abs/1708.06374, 2017.
  [Online]. Available: \url{http://arxiv.org/abs/1708.06374}
\BIBentrySTDinterwordspacing

\bibitem{ITS_2019_workshop}
I.~Alvarez, ``Workshop on automated vehicle safety: Verification, validation
  and transparency,'' in \emph{2019 IEEE Intelligent Transportation Systems
  Conference (ITSC)}.\hskip 1em plus 0.5em minus 0.4em\relax IEEE, 2019.

\bibitem{zhao2016accelerated}
D.~Zhao, H.~Lam, H.~Peng, S.~Bao, D.~J. LeBlanc, K.~Nobukawa, and C.~S. Pan,
  ``Accelerated evaluation of automated vehicles safety in lane-change
  scenarios based on importance sampling techniques,'' \emph{IEEE transactions
  on intelligent transportation systems}, vol.~18, no.~3, pp. 595--607, 2016.

\bibitem{sarkar2019behavior}
A.~Sarkar and K.~Czarnecki, ``A behavior driven approach for sampling rare
  event situations for autonomous vehicles,'' \emph{arXiv preprint
  arXiv:1903.01539}, 2019.

\bibitem{elrofai2016scenario}
H.~Elrofai, D.~Worm, and O.~O. den Camp, ``Scenario identification for
  validation of automated driving functions,'' in \emph{Advanced Microsystems
  for Automotive Applications 2016}.\hskip 1em plus 0.5em minus 0.4em\relax
  Springer, 2016, pp. 153--163.

\bibitem{kalra2016driving}
N.~Kalra and S.~M. Paddock, ``Driving to safety: How many miles of driving
  would it take to demonstrate autonomous vehicle reliability?''
  \emph{Transportation Research Part A: Policy and Practice}, vol.~94, pp.
  182--193, 2016.

\bibitem{menzel2018scenarios}
T.~Menzel, G.~Bagschik, and M.~Maurer, ``Scenarios for development, test and
  validation of automated vehicles,'' in \emph{2018 IEEE Intelligent Vehicles
  Symposium (IV)}.\hskip 1em plus 0.5em minus 0.4em\relax IEEE, 2018, pp.
  1821--1827.

\bibitem{amersbach2019functional}
C.~Amersbach and H.~Winner, ``Functional decomposition—a contribution to
  overcome the parameter space explosion during validation of highly automated
  driving,'' \emph{Traffic injury prevention}, vol.~20, no. sup1, pp. S52--S57,
  2019.

\bibitem{de2017assessment}
E.~de~Gelder and J.-P. Paardekooper, ``Assessment of automated driving systems
  using real-life scenarios,'' in \emph{2017 IEEE Intelligent Vehicles
  Symposium (IV)}.\hskip 1em plus 0.5em minus 0.4em\relax IEEE, 2017, pp.
  589--594.

\bibitem{enable2016enable}
E.-S. Consortium \emph{et~al.}, ``Enable-s3 european project,'' \emph{available
  at: www. enable-s3. eu/(accessed 22 January 2018)}, 2016.

\bibitem{putz2017system}
A.~P{\"u}tz, A.~Zlocki, J.~Bock, and L.~Eckstein, ``System validation of highly
  automated vehicles with a database of relevant traffic scenarios,''
  \emph{situations}, vol.~1, pp. 19--22, 2017.

\bibitem{tuncali2019rapidly}
C.~E. Tuncali and G.~Fainekos, ``Rapidly-exploring random trees for testing
  automated vehicles,'' in \emph{2019 IEEE Intelligent Transportation Systems
  Conference (ITSC)}.\hskip 1em plus 0.5em minus 0.4em\relax IEEE, 2019, pp.
  661--666.

\bibitem{nas}
B.~Zoph and Q.~V. Le, ``Neural architecture search with reinforcement
  learning,'' \emph{arXiv preprint arXiv:1611.01578}, 2016.

\bibitem{rss}
Mobileye, ``Implementing the rss model on nhtsa pre-crash scenarios,'' Tech.
  Rep., 2018.

\bibitem{tuncali2018simulation}
C.~E. Tuncali, G.~Fainekos, H.~Ito, and J.~Kapinski, ``Simulation-based
  adversarial test generation for autonomous vehicles with machine learning
  components,'' in \emph{2018 IEEE Intelligent Vehicles Symposium (IV)}.\hskip
  1em plus 0.5em minus 0.4em\relax IEEE, 2018, pp. 1555--1562.

\bibitem{karunakaran2020efficient}
D.~Karunakaran, S.~Worrall, and E.~Nebot, ``Efficient statistical validation
  with edge cases to evaluate highly automated vehicles,'' \emph{arXiv preprint
  arXiv:2003.01886}, 2020.

\bibitem{koren2018adaptive}
M.~Koren, S.~Alsaif, R.~Lee, and M.~J. Kochenderfer, ``Adaptive stress testing
  for autonomous vehicles,'' in \emph{2018 IEEE Intelligent Vehicles Symposium
  (IV)}.\hskip 1em plus 0.5em minus 0.4em\relax IEEE, 2018, pp. 1--7.

\bibitem{iso26262}
\BIBentryALTinterwordspacing
{International Standardization Organization(ISO)}, ``{ISO 26262-1:2018},''
  2018. [Online]. Available: \url{https://www.iso.org/standard/68383.html}
\BIBentrySTDinterwordspacing

\bibitem{kirovskii1}
O.~Kirovskii and V.~Gorelov, ``Driver assistance systems: analysis, tests and
  the safety case. iso 26262 and iso pas 21448,'' in \emph{IOP Conference
  Series: Materials Science and Engineering}, vol. 534, no.~1.\hskip 1em plus
  0.5em minus 0.4em\relax IOP Publishing, 2019, p. 012019.

\bibitem{iso21448}
\BIBentryALTinterwordspacing
{International Standardization Organization(ISO)}, ``{Road vehicles — Safety
  of the intended functionality},'' 2019. [Online]. Available:
  \url{https://www.iso.org/standard/70939.html}
\BIBentrySTDinterwordspacing

\bibitem{aptiv}
Aptiv, ``Safety first for automated driving,'' Tech. Rep., 2019.

\bibitem{koopman2019safety}
P.~Koopman, U.~Ferrell, F.~Fratrik, and M.~Wagner, ``A safety standard approach
  for fully autonomous vehicles,'' in \emph{International Conference on
  Computer Safety, Reliability, and Security}.\hskip 1em plus 0.5em minus
  0.4em\relax Springer, 2019, pp. 326--332.

\bibitem{koopman2016challenges}
P.~Koopman and M.~Wagner, ``Challenges in autonomous vehicle testing and
  validation,'' \emph{SAE International Journal of Transportation Safety},
  vol.~4, no.~1, pp. 15--24, 2016.

\bibitem{euro2013euro}
N.~Euro, ``Euro ncap test protocol--aeb systems,'' \emph{no. July}, 2013.

\bibitem{peng2012evaluation}
H.~Peng and D.~Leblanc, ``Evaluation of the performance and safety of automated
  vehicles,'' \emph{White Pap. NSF Transp. CPS Work}, 2012.

\bibitem{erdogan2018parametrized}
A.~Erdogan, E.~Kaplan, A.~Leitner, and M.~Nager, ``Parametrized end-to-end
  scenario generation architecture for autonomous vehicles,'' in \emph{2018 6th
  International Conference on Control Engineering \& Information Technology
  (CEIT)}.\hskip 1em plus 0.5em minus 0.4em\relax IEEE, 2018, pp. 1--6.

\bibitem{weber2020simulation}
N.~Weber, D.~Frerichs, and U.~Eberle, ``A simulation-based, statistical
  approach for the derivation of concrete scenarios for the release of highly
  automated driving functions,'' in \emph{AmE 2020-Automotive meets
  Electronics; 11th GMM-Symposium}.\hskip 1em plus 0.5em minus 0.4em\relax VDE,
  2020, pp. 1--6.

\bibitem{amersbach2019defining}
C.~Amersbach and H.~Winner, ``Defining required and feasible test coverage for
  scenario-based validation of highly automated vehicles,'' in \emph{2019 IEEE
  Intelligent Transportation Systems Conference (ITSC)}.\hskip 1em plus 0.5em
  minus 0.4em\relax IEEE, 2019, pp. 425--430.

\bibitem{elrofai2018scenario}
H.~Elrofai, J.-P. Paardekooper, E.~de~Gelder, S.~Kalisvaart, and O.~O. den
  Camp, ``Scenario-based safety validation of connected and automated
  driving,'' \emph{Netherlands Organization for Applied Scientific Research,
  TNO, Tech. Rep}, 2018.

\bibitem{waymo2017road}
L.~Waymo, ``On the road to fully self-driving,'' \emph{Waymo Safety Report},
  pp. 1--43, 2017.

\bibitem{sutton2018reinforcement}
R.~S. Sutton and A.~G. Barto, \emph{Reinforcement learning: An
  introduction}.\hskip 1em plus 0.5em minus 0.4em\relax MIT press, 2018.

\bibitem{van2012reinforcement}
M.~Van~Otterlo and M.~Wiering, ``Reinforcement learning and markov decision
  processes,'' in \emph{Reinforcement Learning}.\hskip 1em plus 0.5em minus
  0.4em\relax Springer, 2012, pp. 3--42.

\bibitem{levine2017cs}
S.~Levine, ``Cs 294: Deep reinforcement learning,'' 2017.

\bibitem{carla_simulator}
A.~Dosovitskiy, G.~Ros, F.~Codevilla, A.~Lopez, and V.~Koltun, ``{CARLA}: {An}
  open urban driving simulator,'' in \emph{Proceedings of the 1st Annual
  Conference on Robot Learning}, 2017, pp. 1--16.

\bibitem{chandra2013speed}
S.~Chandra and A.~K. Bharti, ``Speed distribution curves for pedestrians during
  walking and crossing,'' \emph{Procedia-Social and Behavioral Sciences}, vol.
  104, pp. 660--667, 2013.

\bibitem{liu2017microscopic}
M.~Liu, W.~Zeng, P.~Chen, and X.~Wu, ``A microscopic simulation model for
  pedestrian-pedestrian and pedestrian-vehicle interactions at crosswalks,''
  \emph{PLoS one}, vol.~12, no.~7, p. e0180992, 2017.

\end{thebibliography}

\end{document}